\documentclass[letterpaper]{article} 
\usepackage[preprint]{aaai2027}  
\usepackage[hyphens]{url}  
\usepackage{graphicx} 
\usepackage{natbib}  
\usepackage{caption} 
\usepackage{booktabs}
\usepackage{multirow}
\usepackage{algorithm}
\usepackage{algorithmic}
\usepackage{amsmath}
\usepackage{amssymb}
\title{Beyond Downstream QA: Rethinking How We Evaluate Agent Memory}
\author{
    Sebastian-Antonio Toma\textsuperscript{\rm 1,\rm 2}\corresponding,
    Jean de Kernier\textsuperscript{\rm 1},
    Mike Ply\textsuperscript{\rm 1},
    Oliviu Matei\textsuperscript{\rm 2}
}
\affiliations{
    \textsuperscript{\rm 1}Amazon Web Services\\
    \textsuperscript{\rm 2}Technical University of Cluj-Napoca, Romania\\
    Sebastian.Toma@campus.utcluj.ro
}

\begin{document}

\maketitle

\begin{abstract}
Agent memory benchmarks evaluate whether an agent's responses reflect correct knowledge, measuring downstream question-answering accuracy as a proxy for memory quality. We show this evaluation paradigm is incomplete: it conflates memory correctness with retrieval and generation quality, producing both false positives where correct responses mask broken memory and false negatives where incorrect responses blame correct memory for retrieval failures. We introduce FEB-30, a benchmark of 30 fact-evolution trajectories spanning 6 diagnostic categories and 3 difficulty levels, constructed via paradigm inversion: documents are generated from known fact trajectories, yielding deterministic ground truth without human annotation. Each trajectory defines how a fact should evolve across multi-author artifacts (documents, emails, conversations), enabling direct inspection of the memory store after each ingestion step. Applying FEB-30 to a representative memory system, we demonstrate that intrinsic and extrinsic evaluation diverge by an average of 16 percentage points, with 44\% of trajectories exceeding meaningful divergence thresholds. Diagnostic decomposition reveals that the system correctly classifies memory operations (create, update, reinforce, supersede) only 20\% of the time, a failure entirely invisible to extrinsic evaluation, and that failure modes differ systematically by category and difficulty. We release the benchmark, document generation pipeline, and scoring framework to support reproducible evaluation of agent memory systems.\footnote{\url{https://github.com/amazon-science/feb-30-benchmark}}
\end{abstract}

\section{Introduction}

Large language model (LLM) agents operating over extended time horizons require persistent external memory to accumulate knowledge beyond the ephemeral context window~\cite{coala2024,anatomy2026}. As these systems grow more complex, with over a dozen distinct architectures published since 2023~\cite{latimer2025hindsight,packer2023memgpt,rasmussen2025zep,chhikara2025mem0,ariadnemem2026}, the question of \emph{how to evaluate them} becomes pressing.

A survey of current benchmarks reveals a systematic limitation: every existing evaluation measures memory quality \emph{extrinsically}, by posing questions to the agent and grading its responses~\cite{locomo2024,longmemeval2026,memoryagentbench2026,memora2026,statebench2026,meme2026}. This produces a single composite score that conflates memory, retrieval, and generation quality. When the system fails, one cannot identify which component is responsible; when it succeeds, one cannot confirm the memory is actually correct~\cite{anonymous2026intrinsic}.

We identify two concrete failure modes that extrinsic evaluation cannot detect:

\begin{itemize}
\item \textbf{False positives (LLM compensation):} A strong generation model can produce correct responses despite incorrect or incomplete memory state, by reasoning over partial retrieval results at inference time. Extrinsic evaluation scores this as success.
\item \textbf{False negatives (retrieval bottleneck):} A memory system may correctly store and update facts, yet fail to retrieve them for a specific query due to embedding mismatch or retrieval cutoff. Extrinsic evaluation blames the memory system for a retrieval failure.
\end{itemize}

We propose \emph{intrinsic evaluation}, direct comparison of the memory store contents against known ground truth at each stage of knowledge evolution, as a complementary evaluation paradigm. The relationship mirrors unit versus end-to-end testing in software engineering: intrinsic evaluation checks the memory component's internal state directly, while extrinsic evaluation only observes end-to-end behavior, so a passing end-to-end test can mask a broken component just as a correct component can fail end-to-end due to an integration bottleneck. A parallel from NLP history motivates the same shift: in the 2010s, word embeddings were evaluated only through downstream task performance until intrinsic evaluation (analogy tasks, similarity correlations) enabled faster progress by isolating representation quality~\cite{baroni2014predict,schnabel2015evaluation}.
Our contributions are:
\begin{enumerate}
\item \textbf{FEB-30}, a benchmark of 30 fact-evolution trajectories with deterministic ground truth, spanning 6 diagnostic categories that isolate distinct failure modes invisible to extrinsic evaluation. The benchmark comprises:
\begin{itemize}
\item Fact-evolution trajectories with atomic claim ground truth at each state.
\item A paradigm-inversion document generation pipeline using multi-model rotation and 3-judge verification.
\item A scoring framework measuring both intrinsic (claim coverage, update detection, false update rate) and extrinsic (QA accuracy) metrics.
\end{itemize}
\item \textbf{Intrinsic evaluation methodology} for agent memory: direct inspection of memory state against known ground truth at each stage of knowledge evolution, enabling diagnostic decomposition of failures by pipeline stage.
\item Empirical evidence that intrinsic and extrinsic evaluation \textbf{diverge by 16 percentage points on average}, with 44\% of trajectories exceeding meaningful divergence and failure modes differing systematically by category.
\end{enumerate}

\section{Related Work}

\paragraph{Agent Memory Benchmarks.}
MemoryAgentBench~\cite{memoryagentbench2026} defines four memory competencies (retention, temporal reasoning, knowledge updates, and information integration) evaluated through QA interactions. Memora~\cite{memora2026} introduces the FAMA metric that penalizes reliance on invalidated memories, decomposing performance into freshness-aware components. LongMemEval-V2~\cite{longmemeval2026} tests long-term memory along five temporal reasoning capabilities. AMA-Bench~\cite{amabench2026} evaluates long-horizon memory for agentic applications. STATE-Bench~\cite{statebench2026} measures whether agents improve at tasks through accumulated experience. MEME~\cite{meme2026} introduces controlled state trajectories with multi-entity evolution, the closest methodology to ours, though still evaluated through the response layer.

All these benchmarks share a common structure: pose queries to the agent, grade responses against expected answers. None directly inspects the memory store.

\paragraph{Memory Architecture Surveys.}
Recent surveys~\cite{anatomy2026,du2026memory,luo2026storage,hu2025memoryage,huang2026rethinking} catalog memory architectures and operations. Du et al.~\cite{memoryops2025} define six core memory operations (consolidation, updating, indexing, forgetting, retrieval, condensation), decisions that should be evaluated at the operation level, not only through downstream responses. The Anatomy of Agentic Memory survey~\cite{anatomy2026} identifies benchmark saturation as a key concern, yet proposes more extrinsic benchmarks as the solution.

\paragraph{Conflict Detection Benchmarks.}
ConflictBank~\cite{conflictbank2024} benchmarks LLM behavior under knowledge conflicts but evaluates at query time, not at the memory consolidation stage. MAGIC~\cite{magic2025} tests multi-hop conflict detection in retrieval-augmented generation. These address a related but distinct problem: they test whether a model can \emph{reason} about conflicts in retrieved context, not whether a memory system correctly \emph{detects and resolves} conflicts during ingestion.

\paragraph{Paradigm Inversion.}
RIKER~\cite{riker2025} introduced the concept of generating evaluation corpora from known ground truth for RAG evaluation, avoiding the expense and subjectivity of human annotation on real documents. We extend this approach to temporal fact evolution, generating multi-document sequences where ground truth is a trajectory of states rather than a static set of facts.

\paragraph{Intrinsic vs.\ Extrinsic Evaluation.}
The distinction between intrinsic and extrinsic evaluation is well-established in NLP for embeddings~\cite{baroni2014predict,schnabel2015evaluation} and parsing. The Future AGI practitioner framework~\cite{futureagi2026} explicitly argues that direct memory store inspection is an anti-pattern. A concurrent survey~\cite{anonymous2026intrinsic} documents this gap systematically across 15 benchmarks, arguing for complementary intrinsic evaluation. We present the first benchmark and empirical results that validate this argument.

\section{Benchmark Design: FEB-30}

FEB-30 (Fact Evolution Benchmark, 30 trajectories) is organized along diagnostic categories and difficulty levels (Section~\ref{sec:distribution}). The name is a deliberate analogy: February 30 is a date that does not exist, the calendar approaches it asymptotically (February 28, sometimes 29) but never reaches 30. Similarly, recent benchmarks and surveys have moved progressively closer to intrinsic evaluation, introducing memory-specific tasks~\cite{memoryagentbench2026}, decomposing scores into memory-aware components~\cite{memora2026}, and defining controlled state trajectories~\cite{meme2026}, yet remain on the extrinsic side of the threshold, never directly inspecting the memory store, as illustrated by~\cite{anonymous2026intrinsic}. FEB-30 crosses it.

\subsection{Diagnostic Categories}

Each category targets a specific gap in extrinsic-only evaluation (Table~\ref{tab:categories}):

\begin{table}[t]
\centering
\small
\begin{tabular}{@{}p{0.6cm}p{2.4cm}p{3.9cm}@{}}
\toprule
\textbf{Cat} & \textbf{Name} & \textbf{Design constraint} \\
\midrule
C1 & LLM Compensation & Memory wrong in a way QA still works \\
C2 & Retrieval Bottleneck & Memory correct but hard to retrieve \\
C3 & Classification & Focus on create/update/reinforce/conflict decision correctness \\
C4 & Synthesis Quality & Multi-observation $\rightarrow$ correct fact formation \\
C5 & Temporal Correctness & Fact must evolve correctly over time \\
C6 & Multi-Author & Artifacts from different author roles \\
\bottomrule
\end{tabular}
\caption{FEB-30 diagnostic categories with design constraints. C1 and C2 are designed to produce intrinsic-extrinsic divergence; C3--C6 test dimensions where intrinsic evaluation provides richer signal than extrinsic.}
\label{tab:categories}
\end{table}

\textbf{C1 (LLM Compensation)} constructs scenarios where memory state is incorrect (e.g., outdated fact not updated, contradictions unresolved) but a strong LLM still produces correct answers by reasoning over retrieved fragments, so a system that never updates its facts scores well extrinsically.

\textbf{C2 (Retrieval Bottleneck)} constructs scenarios where facts are correctly stored but queries are phrased to produce low embedding similarity (vocabulary mismatch, different abstraction level). The memory is correct, only retrieval fails.

\textbf{C3 (Classification Accuracy)} tests whether the system correctly classifies each observation as a create, update, reinforce, or conflict operation, decisions extrinsic evaluation cannot observe directly.

\textbf{C4 (Synthesis Quality)} tests whether multiple observations from different artifacts merge into correct, well-formed facts, with ground truth specifying which should form a single fact versus separate facts.

\textbf{C5 (Temporal Correctness)} tests whether the memory state correctly reflects temporal evolution: linear progression (A$\rightarrow$B$\rightarrow$C), retraction, and gradual drift across artifacts.

\textbf{C6 (Multi-Author Attribution)} tests whether source identity affects consolidation correctly: corroboration from different authors should raise confidence differently than repetition from one, and disagreement should trigger conflict detection.

\subsection{Difficulty Levels}

\begin{table}[t]
\centering
\small
\begin{tabular}{@{}lccl@{}}
\toprule
\textbf{Level} & \textbf{Docs} & \textbf{States} & \textbf{Signal clarity} \\
\midrule
L1 & 1--2 & 2--3 & Claims nearly explicit \\
L2 & 3--4 & 3--4 & Claims embedded in context \\
L3 & 4--5 & 4--5 & Requires inference \\
\bottomrule
\end{tabular}
\caption{Difficulty levels in FEB-30.}
\label{tab:difficulty}
\end{table}

At L1 (Table~\ref{tab:difficulty}), target claims appear almost verbatim in generated documents with minimal surrounding noise. At L2, claims are embedded within realistic document context and require extraction from longer passages. At L3, claims must be inferred from combinations of statements, and documents contain substantial unrelated content that could mislead extraction systems.

\subsection{Trajectory Schema}

Each trajectory is defined as a JSON structure specifying:

\begin{itemize}
\item A unique trajectory ID, category, difficulty, and domain assignment.
\item A \textbf{type} indicating the evolution pattern (linear-evolution, contradiction, scope-expansion, retraction, stable, or gradual-drift).
\item A sequence of \textbf{states}, each containing a set of atomic \textbf{claims} representing the expected memory contents after processing documents up to that state.
\item Per-state \textbf{docClaims}, the subset of claims that the document for that state should contain (new information only, not cumulative repetition).
\item An expected \textbf{action} per state transition (create, update, reinforce, conflict, or supersede).
\item \textbf{Author role} and document type metadata for multi-author trajectories.
\end{itemize}

Claims are required to be \emph{atomic} (one assertable proposition per claim) and \emph{self-contained} (understandable without reference to other claims). This ensures scoring is unambiguous: each claim either is or is not entailed by the system's fact store.

The key design choice is \textbf{docClaims vs.\ claims}: documents contain only new information per state. The system must retain prior knowledge across document ingestions, testing actual memory persistence, not single-document extraction.

\subsection{Distribution}
\label{sec:distribution}

FEB-30 contains 30 trajectories across 6 categories, distributed across difficulty levels (10 at L1, 10 at L2, 10 at L3). Trajectories span 6 technical domains (infrastructure, security, financial operations, machine learning, data engineering, and containers) to avoid domain-specific bias. These trajectories instantiate the benchmark in a technical and business-operations setting, but the methodology is domain-agnostic as the same paradigm-inversion construction can produce trajectories for any domain in which facts evolve across multi-author artifacts, by substituting the domain's own topics and observation categories. The total corpus comprises 147 generated documents: 127 trajectory state documents and 20 noise documents (4 per C2 trajectory to test retrieval under distraction).

\section{Document Generation Pipeline}

We adopt paradigm inversion~\cite{riker2025}: rather than annotating ground truth on real documents (expensive and subjective), we generate documents from known fact trajectories, making the trajectory itself the ground truth. Algorithm~\ref{alg:docgen} formalizes the pipeline.

\begin{algorithm}[t]
\caption{Document Generation with Verification}
\label{alg:docgen}
\textbf{Input}: Trajectory $T$ with states $\{s_0, \ldots, s_n\}$\\
\quad Each state $s_i$ has claims $C_i$ (expected memory state, used by scorer) and docClaims $\hat{C}_i$ (new info for this document)\\
\textbf{Input}: Generator pool $\mathcal{M} = \{m_1, \ldots, m_6\}$, Judge pool $\mathcal{J} = \{j_1, j_2, j_3\}$\\
\textbf{Input}: Style references $\mathcal{R}_{\text{AMI}}, \mathcal{R}_{\text{Enron}}$\\
\textbf{Output}: Verified document set $D_T$
\begin{algorithmic}[1]
\FOR{each state $s_i \in T$}
  \STATE $m \leftarrow \mathcal{M}[i \bmod |\mathcal{M}|]$ \hfill \textit{// rotate model}
  \STATE $r \leftarrow \textsc{SelectStyle}(\textsc{type}(s_i), \textsc{diff}(T), \mathcal{R})$
  \FOR{$k = 1$ \textbf{to} $3$}
    \IF{$\textsc{diff}(T) = \text{L3}$}
      \STATE $h \leftarrow \textsc{SelectHost}(\textsc{domain}(T), \mathcal{R})$
      \STATE $d_k \leftarrow \textsc{Inject}(m, \hat{C}_i, h)$
    \ELSE
      \STATE $d_k \leftarrow \textsc{Generate}(m, \hat{C}_i, r, \textsc{diff}(T))$
    \ENDIF
    \STATE $V \leftarrow \bigwedge_{j \in \mathcal{J}} \left[ \textsc{Entail}(j, d_k, \hat{C}_i) \wedge \lnot\textsc{Halluc}(j, d_k, T) \right]$
    \IF{$V = \textit{true}$}
      \STATE $D_T \leftarrow D_T \cup \{d_k\}$; \textbf{break}
    \ELSE
      \STATE $m \leftarrow \mathcal{M}[(i + k) \bmod |\mathcal{M}|]$
    \ENDIF
  \ENDFOR
\ENDFOR
\STATE $D_T \leftarrow D_T \cup \textsc{GenerateNoise}(\mathcal{M}, \textsc{cat}(T), \textsc{domain}(T))$
\STATE \textbf{Constraint:} $\forall m \in \mathcal{M}: |\{d \in D_T : \textsc{gen}(d) = m\}| / |D_T| \leq 0.3$
\RETURN $D_T$
\end{algorithmic}
\end{algorithm}

\subsection{Multi-Model Rotation}

Documents are generated using 6 models across 5 providers, rotated per state index within each trajectory to simulate different human authors with different writing styles. No single model generates more than 30\% of total documents. Following evidence that multi-source synthetic data mitigates distribution collapse~\cite{multimodel2026} and reduces single-generator bias~\cite{syntheticeval2025}, this additionally simulates the natural stylistic variation of multi-author corpora.

The generation models are all different from the system's extraction model (which uses a different LLM family), avoiding the stylistic bias that inflates scores when the same model generates and extracts~\cite{syntheticeval2025}.

\subsection{Style-Grounded Generation}

Documents follow a realism gradient across difficulty levels:

\begin{enumerate}
\item \textbf{Style-grounded (L1/L2):} An LLM generates a new document for the trajectory claims but mimics the tone, vocabulary, and structure of a real reference document.
\item \textbf{Hybrid injection (L3):} Claims are inserted into actual human-written documents from unrelated domains, producing text that is approximately 90\% authentic.
\end{enumerate}

For style-grounded generation, documents are typed (meeting notes, email, architecture review, status update, executive briefing) and matched to style references from a curated set of 12 AMI meeting summaries~\cite{ami_corpus} and 12 Enron emails~\cite{enron} selected for diversity in length, formality, and structure.

Document length and noise scale with difficulty:

\begin{itemize}
\item \textbf{L1:} 200--400 words, no noise claims, target claims nearly explicit.
\item \textbf{L2:} 400--800 words, 2--3 unrelated noise claims, target claims embedded in context.
\item \textbf{L3:} 600--1000 words, 4--5 noise claims, subtle and ambiguous signal.
\end{itemize}

For L3 trajectories, host documents are selected from unrelated domains (e.g., product design meetings, energy pipeline discussions) to ensure no semantic overlap with the trajectory topic, making scoring unambiguous: any relevant fact the system extracts must come from the injected section.

\subsection{Verification Pipeline}

The verification criterion (Algorithm~\ref{alg:docgen}, line 6) requires unanimous agreement from 3 independent judge models (different from the generators), following the multi-judge grounding methodology of~\cite{memora2026} and~\cite{riker2025}. Each judge assesses: (1) all target claims are inferable from the document, (2) no extra factual information beyond the trajectory is introduced, and (3) the document reads naturally.

Additionally, documents underwent human review, with 11\% requiring content corrections (forward leakage, narrative contradictions, cross-document inconsistencies, numeric mismatches) and 28\% requiring scaffolding removal (code fences, LLM preamble such as ``generated email about...'' or ``injected claims into...'') that does not affect extraction.

\section{Evaluation Framework}

We evaluate each system using both intrinsic and extrinsic metrics on the same trajectories, enabling direct measurement of their divergence.

\subsection{Intrinsic Metrics}

After processing all artifacts for a trajectory $T$, we compare the system's fact store $F$ against the expected claims $C_i$ at each state $s_i$. Let $\text{Entail}(f, c) = 1$ if fact $f$ entails claim $c$ (determined by embedding similarity $> 0.8$ followed by LLM judge verification).

\textbf{Claim coverage} at state $s_i$:
\begin{equation}
\text{Cov}(s_i) = \frac{|\{c \in C_i : \exists f \in F,\ \text{Entail}(f, c) = 1\}|}{|C_i|}
\end{equation}

\textbf{Fact accuracy} across the trajectory:
\begin{equation}
\text{Acc}(T) = \frac{1}{|T|} \sum_{i=0}^{|T|-1} \text{Cov}(s_i)
\end{equation}

\textbf{Update detection recall}, where $U = \{s_i : a_i \in \{\text{update}, \text{supersede}\}\}$ is the set of states requiring a change:
\begin{equation}
\text{UDR}(T) = \frac{|\{s_i \in U : \text{changed}(F, s_{i-1}, s_i)\}|}{|U|}
\end{equation}

\textbf{False update rate}, where $R = \{s_i : a_i = \text{reinforce}\}$:
\begin{equation}
\text{FUR}(T) = \frac{|\{s_i \in R : \text{changed}(F, s_{i-1}, s_i)\}|}{|R|}
\end{equation}

Additionally, we report \textbf{fact count} (number of active facts per trajectory, detecting duplication or loss) and \textbf{confidence MAE} ($\frac{1}{|T|}\sum_i |c_{\text{expected}}(s_i) - c_{\text{actual}}(s_i)|$).

\subsection{Extrinsic Metrics}

For each trajectory, we generate diagnostic questions whose answers depend on the final memory state. The system retrieves relevant facts via its standard retrieval mechanism, then an LLM generates an answer. A judge model scores each answer as YES (1.0), PARTIAL (0.5), or NO (0.0) against the expected answer derived from ground-truth claims.

\subsection{Divergence Metric}

The key metric is \textbf{mean absolute divergence} (MAD) between intrinsic and extrinsic scores:
\begin{equation}
\text{MAD} = \frac{1}{N} \sum_{T} |\text{Acc}(T) - \text{Extr}(T)|
\end{equation}

MAD quantifies how much information is lost by relying on extrinsic evaluation alone, without requiring an arbitrary threshold. As a secondary interpretable metric, we report the \textbf{divergence rate}: the percentage of trajectories where the two evaluations disagree by more than $\tau = 0.15$:
\begin{equation}
\text{DivRate} = \frac{1}{N} \sum_{T} \mathbb{1}\left[|\text{Acc}(T) - \text{Extr}(T)| > \tau\right]
\end{equation}

\section{Experiments}

\subsection{System Under Test}

\begin{figure*}[t]
\centering
\setlength{\unitlength}{1cm}
\begin{picture}(16,6.2)

\put(0.0,5.2){\scriptsize\textbf{Ingestion Path (per state $s_i$ of trajectory $T$)}}

\put(0.2,3.8){\framebox(2.0,0.9){\parbox{1.8cm}{\centering\small Artifact\\Ingestion}}}
\put(2.8,3.8){\framebox(2.2,0.9){\parbox{2cm}{\centering\small Observation\\Extraction}}}

\put(5.6,2.8){\framebox(8.0,2.6){}}
\put(5.8,5.0){\scriptsize\textbf{Fact Synthesis (per observation)}}

\put(6.0,3.8){\framebox(2.0,0.9){\parbox{1.8cm}{\centering\small Embed +\\Retrieve}}}
\put(8.6,3.8){\framebox(2.0,0.9){\parbox{1.8cm}{\centering\small LLM\\Classify}}}
\put(11.2,3.8){\framebox(2.0,0.9){\parbox{1.8cm}{\centering\small Apply\\Action}}}

\put(14.2,3.4){\framebox(1.6,1.2){\parbox{1.4cm}{\centering\small Fact\\Store}}}

\put(2.2,4.25){\vector(1,0){0.6}}
\put(5.0,4.25){\vector(1,0){0.6}}
\put(8.0,4.25){\vector(1,0){0.6}}
\put(10.6,4.25){\vector(1,0){0.6}}

\put(13.6,4.05){\vector(1,0){0.6}}

\put(14.2,3.6){\vector(-1,0){0.6}}

\put(8.8,3.0){\scriptsize\textit{create $|$ reinforce $|$ update $|$ supersede $|$ none}}
\put(10.9,3.2){\line(0,1){0.6}}

\put(0.0,2.6){\scriptsize\textbf{$\star$ Intrinsic Scoring (per state $s_i$)}}

\put(0.2,1.4){\framebox(2.0,0.9){\parbox{1.8cm}{\centering\small Load Ground\\Truth $C_i$}}}
\put(3.0,1.4){\framebox(2.2,0.9){\parbox{2.0cm}{\centering\small Embed +\\Retrieve}}}
\put(5.8,1.4){\framebox(2.2,0.9){\parbox{2.0cm}{\centering\small LLM Entail-\\ment Judge}}}
\put(8.6,1.4){\framebox(2.2,0.9){\parbox{2.0cm}{\centering\small Compute\\Cov$(s_i)$}}}

\put(2.2,1.85){\vector(1,0){0.8}}
\put(5.2,1.85){\vector(1,0){0.6}}
\put(8.0,1.85){\vector(1,0){0.6}}

\put(14.6,3.4){\line(0,-1){0.8}}
\put(14.6,2.6){\line(-1,0){10.5}}
\put(4.1,2.6){\vector(0,-1){0.3}}

\put(0.0,1.1){\scriptsize\textbf{Extrinsic Scoring (per trajectory $T$)}}

\put(0.2,0.0){\framebox(2.0,0.9){\parbox{1.8cm}{\centering\small Load\\Question}}}
\put(3.0,0.0){\framebox(2.2,0.9){\parbox{2.0cm}{\centering\small Embed +\\Retrieve}}}
\put(5.8,0.0){\framebox(2.0,0.9){\parbox{1.8cm}{\centering\small LLM\\Answer}}}
\put(8.4,0.0){\framebox(2.2,0.9){\parbox{2.0cm}{\centering\small LLM Judge\\Extr$(T)$}}}

\put(2.2,0.45){\vector(1,0){0.8}}
\put(5.2,0.45){\vector(1,0){0.6}}
\put(7.8,0.45){\vector(1,0){0.6}}

\put(15.4,3.4){\line(0,-1){2.15}}
\put(15.4,1.25){\line(-1,0){11.3}}
\put(4.1,1.25){\vector(0,-1){0.35}}

\end{picture}
\caption{System architecture showing the write path (ingestion and fact synthesis), intrinsic and extrinsic scoring pipelines. For each trajectory state $s_i$, intrinsic evaluation compares ground-truth claims $C_i$ against the fact store via embedding retrieval and LLM entailment. Extrinsic evaluation queries the store and judges the response against expected answers. $\star$ = proposed here.}
\label{fig:pipeline}
\end{figure*}
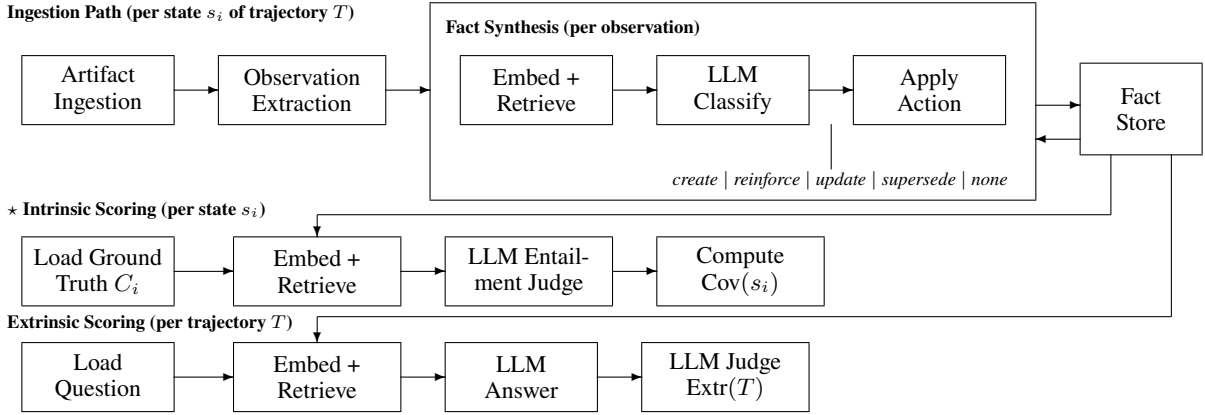

To demonstrate the benchmark's diagnostic utility, we evaluate a representative agent memory system (Figure~\ref{fig:pipeline}). The system follows the dominant architectural pattern shared by virtually all contemporary memory systems, Hindsight~\cite{latimer2025hindsight}, Mem0~\cite{chhikara2025mem0}, AriadneMem~\cite{ariadnemem2026}, and commercial agent memory services alike: (1)~extract structured observations from artifacts (documents, emails, meeting transcripts), (2)~embed them, (3)~retrieve related existing facts via vector search, (4)~classify the relationship via LLM (create, update, reinforce, conflict, or skip), and (5)~synthesize into evolving facts with history. This extract-embed-classify-synthesize pipeline represents the current state of the art; despite surface differences in implementation details (confidence stepping in Hindsight, entropy-aware gating in AriadneMem, Add/Update/No-op in Mem0), the core information flow is shared. The specific system is not the contribution; any system implementing this pattern could be substituted. Our goal is to demonstrate what the benchmark reveals about this class of systems, not to evaluate a particular implementation.

We adopt a two-layer terminology: \emph{observations} are static, atomic factual statements extracted from a single artifact (typed into categories such as decisions, commitments, or milestones), while \emph{facts} are the evolving, synthesized knowledge units that the system maintains over time. Observations are immutable once extracted; facts are the persistent memory that gets created, reinforced, updated, or superseded as new observations arrive. Not all observations aggregate into facts: some are too granular or transient (e.g., a one-time action item), and evaluating observations in isolation would only measure extraction completeness and atomicity, not whether the system forms and maintains correct knowledge. Other systems use different terms for similar concepts (Hindsight calls atomic extractions ``facts'' and synthesized summaries ``observations''; Mem0 uses ``memories'' for both layers without distinction), but the underlying two-layer pattern, extraction followed by aggregation, is shared. Intrinsic evaluation targets the fact layer, as this is the knowledge state that determines system behavior.

Each fact maintains a numeric confidence score (0--1), strengthened by +0.1 per reinforcing observation and adjustable by the LLM on create, update, or supersede decisions. A \emph{temporal guard} prevents stale observations from overwriting newer knowledge: if an observation is older than the matched fact's last update, update and supersede decisions are automatically downgraded to reinforce. Facts track a full chronological history (statement, confidence, change type, and trigger observation at each step), enabling intrinsic evaluation to inspect the complete evolution path. The system additionally tracks the number of distinct contributing authors per fact, counting unique contributors across supporting observations, which enables C6 (multi-author) evaluation. Retrieval uses hybrid search combining BM25~\cite{robertson2009bm25} keyword matching (weight 0.3) with k-NN vector similarity (weight 0.7) over 1024-dimensional embeddings. All LLM classification and synthesis calls use temperature 0 for deterministic outputs; remaining variance across runs comes from the embedding retrieval layer. The system runs on cloud-hosted LLM APIs for both extraction and synthesis (prompts and models in Appendices~\ref{app:prompts}, \ref{app:models}).

Formally, for each new observation $o$ with embedding $e_o$, the system retrieves the most similar existing fact $f = \arg\max_{f' \in F} \text{sim}(e_o, e_{f'})$ and classifies the action:
\begin{equation}
a = \textsc{Classify}_{\text{LLM}}(o, f) \in \{\text{create}, \text{reinforce}, \text{update}, \ldots\}
\end{equation}
A temporal guard enforces recency: $a \leftarrow \text{reinforce}$ if $a \in \{\text{update}, \text{supersede}\} \wedge t_o < t_f^{\text{last}}$. Regardless of action, the observation is linked to the fact: $S_f \leftarrow S_f \cup \{o\}$, where $S_f$ is the set of supporting observations for $f$. Confidence evolves as:
\begin{equation}
c_f^{(t+1)} = \begin{cases}
c_{\text{LLM}}(o) & \text{if } a = \text{create} \\
\min(c_f^{(t)} + 0.1,\; 1) & \text{if } a = \text{reinforce} \\
c_{\text{LLM}}(s_f^{(t+1)}) & \text{if } a \in \{\text{update}, \text{supersede}\} \\
c_f^{(t)} & \text{if } a = \text{none}
\end{cases}
\end{equation}
and the fact statement as:
\begin{equation}
s_f^{(t+1)} = \begin{cases}
\textsc{Init}_{\text{LLM}}(o) & \text{if } a = \text{create} \\
s_f^{(t)} & \text{if } a \in \{\text{reinforce}, \text{none}\} \\
\textsc{Agg}_{\text{LLM}}(s_f^{(t)}, o) & \text{if } a \in \{\text{update}, \text{supersede}\}
\end{cases}
\end{equation}

\subsection{Results}

\begin{table*}[t]
\centering
\small
\begin{minipage}[t]{0.52\textwidth}
\centering
\begin{tabular}{@{}lcccc@{}}
\toprule
\textbf{Category} & Intr. (\%) & Extr. (\%) & MAD & DivRate (\%) \\
\midrule
C1: LLM Comp.    & 82.2{\scriptsize$\pm$2.9}  & 54.5{\scriptsize$\pm$10.8}  & .295{\scriptsize$\pm$.084} & 67.3 \\
C2: Retrieval    & 60.3{\scriptsize$\pm$4.7}  & 50.9{\scriptsize$\pm$2.9}  & .135{\scriptsize$\pm$.034} & 45.5 \\
C3: Classif.     & 70.8{\scriptsize$\pm$8.2}  & 48.9{\scriptsize$\pm$16.4}  & .258{\scriptsize$\pm$.117} & 54.5 \\
C4: Synthesis    & 52.8{\scriptsize$\pm$3.1}  & 49.2{\scriptsize$\pm$2.4}  & .104{\scriptsize$\pm$.036} & 21.2 \\
C5: Temporal     & 59.3{\scriptsize$\pm$5.2}  & 47.3{\scriptsize$\pm$4.5}  & .144{\scriptsize$\pm$.062} & 45.5 \\
C6: Multi-Auth.  & 41.5{\scriptsize$\pm$4.8}  & 41.8{\scriptsize$\pm$5.7}  & .106{\scriptsize$\pm$.049} & 34.5 \\
\midrule
\textbf{Overall} & 60.6{\scriptsize$\pm$1.6}  & 48.8{\scriptsize$\pm$2.9}  & .168{\scriptsize$\pm$.025} & 43.6 \\
\bottomrule
\end{tabular}
\end{minipage}%
\hfill
\begin{minipage}[t]{0.44\textwidth}
\centering
\begin{tabular}{@{}lcccc@{}}
\toprule
\textbf{Diff.} & Intr. (\%) & Extr. (\%) & MAD & DivRate (\%) \\
\midrule
L1 & 68.0{\scriptsize$\pm$3.2} & 50.9{\scriptsize$\pm$4.2} & .178{\scriptsize$\pm$.051} & 54.5 \\
L2 & 67.0{\scriptsize$\pm$4.5} & 47.7{\scriptsize$\pm$7.8} & .224{\scriptsize$\pm$.071} & 50.0 \\
L3 & 46.7{\scriptsize$\pm$1.6} & 47.7{\scriptsize$\pm$3.3} & .103{\scriptsize$\pm$.045} & 26.4 \\
\midrule
 & 60.6{\scriptsize$\pm$1.6} & 48.8{\scriptsize$\pm$2.9} & .168{\scriptsize$\pm$.025} & 43.6 \\
\bottomrule
\end{tabular}
\end{minipage}
\caption{Intrinsic (fact accuracy \%) and extrinsic (QA accuracy \%) scores by category (left) and difficulty (right), averaged across $N{=}11$ runs ($\pm$ std). MAD = mean absolute divergence between intrinsic and extrinsic per trajectory. DivRate = percentage with divergence $> \tau{=}0.15$. C1 shows highest MAD (.295): memory quality is systematically better than what QA evaluation reveals. Per-trajectory scores are in Appendix~\ref{app:pertraj}.}
\label{tab:results}
\end{table*}

\paragraph{Finding 1: Intrinsic-Extrinsic Divergence.}
Across all 30 trajectories (Table~\ref{tab:results}), the mean absolute divergence between intrinsic and extrinsic evaluation is $0.168 \pm 0.025$ (over 16 percentage points), with 44\% of trajectories exceeding the $\tau = 0.15$ threshold ($N{=}11$ runs). Using a stricter pass/fail criterion ($\geq$0.5 vs.\ $<$0.5), in $21 \pm 4$\% of trajectories one evaluation method passes while the other fails, meaning a practitioner relying on extrinsic evaluation alone would reach the wrong conclusion about system correctness in roughly 1 out of 5 cases (Cohen's $\kappa = 0.118$, barely above chance), confirming the two evaluation methods carry largely independent signal. A Pearson correlation between intrinsic and extrinsic scores across all 330 paired evaluations is $r = 0.423$ ($p < 10^{-15}$), so $r^2 \approx 0.18$: only 18\% of the variance in one score is explained by the other, 82\% independent. The signed mean divergence is $+0.118 \pm 0.029$, indicating intrinsic scores are systematically higher than extrinsic: the system's memory is on average better than what QA evaluation reveals. A Wilcoxon signed-rank test over all 330 paired trajectory evaluations ($W = 4042$, $p < 10^{-16}$) confirms this divergence is statistically significant, establishing that the two evaluation methods measure fundamentally different properties of the system. C1 (LLM Compensation) shows the highest MAD ($0.295 \pm 0.084$): intrinsic scores 82.2\% while extrinsic scores only 54.5\%, confirming that memory quality is systematically underestimated by response-based evaluation. C5 (Temporal) and C6 (Multi-Author) show the lowest intrinsic scores (59.3\% and 41.5\%), identifying temporal evolution and source attribution as the hardest dimensions, yet extrinsic evaluation reports near-uniform scores across all categories, masking these weaknesses.

\paragraph{Finding 2: Difficulty and Divergence.}
DivRate decreases with difficulty (L1: 55\%, L2: 50\%, L3: 26\%). At L1, intrinsic substantially exceeds extrinsic ($68.0 \pm 3.2$ vs.\ $50.9 \pm 4.2$), meaning the system stores correct knowledge that retrieval fails to surface for QA. At L3, both scores converge near 47\%, but diverge on individual trajectories in opposite directions (some false positives, some false negatives). MAD varies across levels (.178, .224, .103), indicating disagreement persists at all difficulty levels. Classification accuracy degrades sharply with difficulty ($33.8 \rightarrow 16.5 \rightarrow 10.1$\%), confirming that the system's ability to correctly label actions deteriorates as signal becomes more subtle.

\paragraph{Finding 3: Category-Specific Failure Modes.}
Diagnostic decomposition reveals distinct failure patterns. The overall false update rate is $22.2 \pm 1.2$\%: the system modifies facts when it should only reinforce. Update detection recall is $73.6 \pm 2.6$\%. Classification accuracy (whether the system correctly labels the action as create/reinforce/update/supersede) is only $20.1 \pm 2.7$\%, with C2 (4.5\%) and C5 (0.4\%) near zero. Confidence MAE is $0.106 \pm 0.002$ across all trajectories. C6 (Multi-Author) shows the lowest intrinsic accuracy (41.5\%), identifying multi-author source attribution as the system's weakest dimension. These insights are invisible to extrinsic evaluation, which reports a near-uniform 42--55\% across most categories (per-category/difficulty breakdowns in Appendix~\ref{app:udrfur}).

\paragraph{Finding 4: Actionable Diagnostics.}
The intrinsic metrics directly identify which pipeline stage to improve: (1) classification accuracy is 20.1\% overall, meaning the system rarely assigns the correct action label even when its fact state ends up partially correct; (2) multi-author source attribution (C6: 41.5\%) does not adequately leverage author identity; (3) false updates (22.2\%) indicate overly aggressive modification on reinforcement signals. None of these are recoverable from extrinsic scores.

\paragraph{Finding 5: Robustness Across Systems.}
To test whether the intrinsic--extrinsic divergence is an artifact of a particular system
implementation, corpus version, or backbone model, we evaluate four configurations spanning two
factors: system configuration (simpler vs.\ main) and corpus (pre-release vs.\ the publicly
released set), plus the main configuration with an upgraded backbone model. The
simpler system on the pre-release corpus gives MAD $= 0.164 \pm 0.027$, DivRate $= 43.3\%$,
signed divergence $+0.088 \pm 0.034$, and $\kappa = 0.104$. Holding the corpus fixed (pre-release)
but upgrading to the main system configuration shifts intrinsic accuracy up most sharply among all
four configurations ($57.6\% \rightarrow 61.1\%$) and raises DivRate to its highest value
($49.4\%$, MAD $= 0.181 \pm 0.022$, signed divergence $+0.126 \pm 0.025$). Switching to the publicly
released corpus on top of the main system configuration (used for all other results in this paper)
pulls DivRate back down to $43.6\%$ (MAD $= 0.168 \pm 0.025$, signed divergence
$+0.118 \pm 0.029$, $\kappa = 0.118$), close to the simpler-system starting point; the system
upgrade and the corpus release thus move DivRate in opposite directions and partially cancel. Finally, re-scoring the main configuration with an upgraded
backbone model gives MAD $= 0.158 \pm 0.035$, DivRate $= 39.7\%$, signed divergence
$+0.121 \pm 0.036$, and $\kappa = 0.035$. Across all four configurations, intrinsic accuracy
remains in a narrow $57.6$--$61.4\%$ band, extrinsic accuracy is essentially flat
($48.5$--$49.2\%$), and signed divergence stays positive throughout; pass/fail agreement stays
near chance in every configuration, falling further under the upgraded backbone, consistent with
weak-to-absent agreement rather than a single stable value. The configuration used for the main results in this paper adds
retrieval relevance filtering, concurrent fact-update handling, and improved large-document
extraction; the simpler-system ablation lacks these refinements, which accounts for its modestly
lower intrinsic accuracy. All configurations implement the same extract-embed-classify-synthesize
pattern (Figure~\ref{fig:pipeline}). The divergence therefore persists across all four,
with full statistics in Appendix~\ref{app:robustness}.

\subsection{Diagnostic Decomposition}

\begin{table*}[t]
\centering
\small
\begin{minipage}[t]{0.52\textwidth}
\centering
\begin{tabular}{@{}lccc@{}}
\toprule
\textbf{Category} & Miss (\%) & Fact Count & ClassAcc (\%) \\
\midrule
C1: LLM Comp.    & 54.0{\scriptsize$\pm$4.5} & 13.6{\scriptsize$\pm$1.0} & 41.7{\scriptsize$\pm$2.8} \\
C2: Retrieval    & 74.5{\scriptsize$\pm$4.2} & 20.8{\scriptsize$\pm$1.3} & 4.5{\scriptsize$\pm$5.7} \\
C3: Classif.     & 60.5{\scriptsize$\pm$7.9} & 14.8{\scriptsize$\pm$1.4} & 30.7{\scriptsize$\pm$12.3} \\
C4: Synthesis    & 72.2{\scriptsize$\pm$2.6} & 14.9{\scriptsize$\pm$1.0} & 17.4{\scriptsize$\pm$2.4} \\
C5: Temporal     & 68.0{\scriptsize$\pm$5.3} & 21.0{\scriptsize$\pm$1.0} & 0.4{\scriptsize$\pm$1.1} \\
C6: Multi-Auth.  & 87.5{\scriptsize$\pm$3.5} & 12.0{\scriptsize$\pm$0.8} & 28.8{\scriptsize$\pm$7.3} \\
\midrule
\textbf{Overall} & 71.4{\scriptsize$\pm$2.1} & 16.6{\scriptsize$\pm$0.4} & 20.1{\scriptsize$\pm$2.7} \\
\bottomrule
\end{tabular}
\end{minipage}%
\hfill
\begin{minipage}[t]{0.44\textwidth}
\centering
\begin{tabular}{@{}lccc@{}}
\toprule
\textbf{Diff.} & Miss (\%) & Fact Count & ClassAcc (\%) \\
\midrule
L1 & 54.4{\scriptsize$\pm$5.1} & 9.6{\scriptsize$\pm$0.9} & 33.8{\scriptsize$\pm$6.8} \\
L2 & 61.1{\scriptsize$\pm$3.2} & 14.7{\scriptsize$\pm$0.7} & 16.5{\scriptsize$\pm$2.7} \\
L3 & 86.3{\scriptsize$\pm$1.6} & 22.4{\scriptsize$\pm$0.8} & 10.1{\scriptsize$\pm$4.3} \\
\midrule
\textbf{Overall} & 71.4{\scriptsize$\pm$2.1} & 16.6{\scriptsize$\pm$0.4} & 20.1{\scriptsize$\pm$2.7} \\
\bottomrule
\end{tabular}
\end{minipage}
\caption{Diagnostic decomposition by category (left) and difficulty (right), averaged across $N{=}11$ runs ($\pm$ std). Miss Rate = percentage of expected claims not found in the fact store. Fact Count = average active facts per trajectory (lower indicates better synthesis). ClassAcc = action classification accuracy. C6 shows highest miss rate (88\%); fact count inflates from 9.6 at L1 to 22.4 at L3.}
\label{tab:decomp}
\end{table*}

A key advantage of intrinsic evaluation is \emph{diagnostic decomposition}: when a system fails, we can identify which pipeline stage is responsible:

\begin{itemize}
\item \textbf{Extraction failures:} Claims present in artifacts but not captured by the system's observation layer.
\item \textbf{Synthesis failures:} Claims extracted but incorrectly merged, duplicated, or lost during fact aggregation.
\item \textbf{Update failures:} State transitions missed or false updates applied.
\item \textbf{Conflict failures:} Contradictions not detected or resolved incorrectly.
\end{itemize}

Table~\ref{tab:decomp} reveals three co-occurring failure patterns that intensify with difficulty. First, the claim miss rate rises from 58\% at L1 to 88\% at L3, indicating that subtle or embedded claims are increasingly lost during extraction. Second, fact count inflates from 9.1 to 20.7, showing the system fragments knowledge into many small, partially redundant facts rather than synthesizing them into coherent evolving beliefs. Third, classification accuracy degrades from 37\% to 10\%, meaning the system rarely identifies the correct action (create vs.\ update vs.\ reinforce) at higher difficulty. These three signals are mutually reinforcing: poor classification leads to creating new facts instead of updating existing ones (inflating count), which in turn fragments claims across facts (increasing miss rate on claim-level matching). None of these diagnostics are available from the extrinsic QA score, which remains flat at $\sim$50\% regardless of the underlying failure pattern.

\section{Discussion and Limitations}

\paragraph{Complementary, Not Replacement.}
Intrinsic evaluation does not replace extrinsic evaluation, it complements it. Extrinsic captures end-to-end utility as intrinsic enables diagnosis. A system scoring high on both is genuinely effective; high extrinsic but low intrinsic signals fragility (dependent on LLM compensation); high intrinsic but low extrinsic indicates a retrieval or generation bottleneck.

\paragraph{Synthetic Document Validity.}
Generated documents may not capture all properties of real multi-author text. We mitigate this through style grounding, multi-model rotation, and hybrid injection at L3. Future work should validate transfer via rank correlation with human-annotated real document sequences.

\paragraph{Scope and Limitations.}
FEB-30 targets factual evolution across multi-author artifacts. It does not evaluate episodic memory, procedural memory, or single-turn conversational recall, capabilities addressed by existing benchmarks. The benchmark contains 30 trajectories, sufficient for diagnostic evaluation but not for fine-grained statistical comparisons. Scoring relies on embedding similarity thresholds and LLM judges, introducing potential measurement noise.

\section{Conclusion}

We introduced FEB-30, the first benchmark designed for intrinsic evaluation of agent memory systems. By constructing trajectories where the expected memory state is known at every step, we demonstrated that intrinsic and extrinsic evaluation diverge by an average of 16 percentage points, with 44\% of trajectories showing meaningful disagreement, revealing false positives and false negatives invisible to response-based benchmarks. Diagnostic decomposition showed the system correctly classifies memory operations only 20\% of the time (degrading from 34\% at L1 to 10\% at L3), a dimension extrinsic evaluation cannot measure.

We argue that intrinsic evaluation should become standard practice alongside extrinsic benchmarks, analogous to how intrinsic evaluation of embeddings complemented downstream task evaluation in NLP. We release FEB-30, the generation pipeline, and the scoring framework for reproducible evaluation of agent memory.

\bibliography{aaai2027}

\appendix

\begin{center}
{\Large\bf Appendix}
\end{center}

\section{Trajectory Examples}

\subsection{c5-linear: Linear Temporal Evolution}

\begin{scriptsize}
\begin{verbatim}
{
  "id": "c5-linear",
  "topic": "Database migration from PostgreSQL
            to Aurora",
  "category": "C5",
  "difficulty": "L1",
  "trajectoryType": "linear-evolution",
  "customer": "Meridian Corp",
  "extrinsicQA": ["Describe Meridian Corp's
    database migration: source, target, timeline,
    status?"],
  "states": [
    {
      "index": 0,
      "timestamp": "2025-01-15",
      "claims": [
        "Customer runs PostgreSQL 14 on-premise",
        "Production database holds approx 3 TB"
      ],
      "action": "create",
      "confidence": 0.9
    },
    {
      "index": 1,
      "timestamp": "2025-03-20",
      "claims": [
        "Customer runs PostgreSQL 14 on-premise",
        "Production database holds approx 3 TB",
        "Evaluating Aurora and AlloyDB"
      ],
      "docClaims": [
        "Evaluating Aurora and AlloyDB",
        "PoC planned for Q2"
      ],
      "action": "update",
      "confidence": 0.8,
      "changeReason": "Architecture review"
    }
  ]
}
\end{verbatim}
\end{scriptsize}

\begin{table}[h]
\centering
\small
\begin{tabular}{@{}clp{5cm}@{}}
\toprule
$s_i$ & Action & Expected claims $C_i$ (cumulative) \\
\midrule
0 & create & ``PostgreSQL 14 on-premise in Frankfurt''; ``3~TB database'' \\
1 & update & +``Evaluating Aurora and AlloyDB'' \\
2 & update & +``Chose Aurora''; ``Cutover Q3 2025'' \\
3 & update & +``DMS replication running''; ``95\% stored procedures validated'' \\
\bottomrule
\end{tabular}
\caption{State evolution for c5-linear.}
\end{table}

Each state's document contains only \emph{docClaims} (new information): the $s_2$ document mentions the Aurora decision and Q3 timeline but does \emph{not} repeat that the database is 3~TB. The system must retain $s_0$ claims from memory.

\textbf{Intrinsic scoring} inspects the fact store after each artifact is ingested. At each state $s_i$, the system should have a fact containing all claims from $C_i$, with history showing the correct transitions. If the system creates separate facts for each state (fragmentation) or drops $s_0$ claims (retention failure), intrinsic scoring detects it at the state where it occurs.

\textbf{Extrinsic scoring} poses the question to an agent that retrieves from the memory system's fact store and generates an answer: ``Describe the database migration: source, target, timeline, and status.'' A strong LLM can answer this correctly even if the fact store is fragmented across 4 separate facts, as long as retrieval surfaces all fragments.

\textbf{Actual result (N=11 runs):} Intrinsic = 40--67\%, Extrinsic = 50\%. The system partially fragments the fact, which intrinsic detects but extrinsic masks.

\subsection{c4-out-of-order: Out-of-Order Ingestion}

This trajectory tests temporal robustness: a newer document (the decision) arrives first, then an older document (the evaluation) arrives second. The system must enrich the existing fact with context from the earlier document without overwriting the confirmed decision.

\begin{scriptsize}
\begin{verbatim}
{
  "id": "c4-out-of-order",
  "topic": "Compute platform decision",
  "category": "C4",
  "difficulty": "L2",
  "trajectoryType": "linear-evolution",
  "customer": "Vertex AI Labs",
  "extrinsicQA": ["What compute platform did
    Vertex AI Labs choose, what alternatives
    were considered, and why?"],
  "states": [
    {
      "index": 0,
      "timestamp": "2025-03-10",
      "claims": [
        "Customer chose ECS Fargate",
        "Decision driven by sub-100ms response"
      ],
      "action": "create",
      "confidence": 0.9
    },
    {
      "index": 1,
      "timestamp": "2025-01-20",
      "claims": [
        "Customer chose ECS Fargate",
        "Decision driven by sub-100ms response",
        "Lambda rejected due to cold starts",
        "EKS rejected as over-complex"
      ],
      "docClaims": [
        "Evaluating Lambda, ECS Fargate, and EKS",
        "Lambda cold starts concern given p99 SLA",
        "EKS too complex for team of 4"
      ],
      "action": "update",
      "confidence": 0.9,
      "changeReason": "Earlier evaluation doc
        enriches fact without overwriting"
    },
    {
      "index": 2,
      "timestamp": "2025-05-15",
      "claims": [
        "Customer chose ECS Fargate",
        "Decision driven by sub-100ms response",
        "Lambda rejected due to cold starts",
        "EKS rejected as over-complex",
        "Running 12 Fargate services at 50K rps"
      ],
      "docClaims": [
        "12 Fargate services at 50K rps peak",
        "34ms avg response, within target"
      ],
      "action": "update",
      "confidence": 0.95,
      "changeReason": "Operational review confirms"
    }
  ]
}
\end{verbatim}
\end{scriptsize}

\begin{table}[h]
\centering
\small
\begin{tabular}{@{}cclp{4.5cm}@{}}
\toprule
$s_i$ & Timestamp & Action & Expected claims $C_i$ (cumulative) \\
\midrule
0 & 2025-03-10 & create & ``Chose ECS Fargate''; ``Sub-100ms response'' \\
1 & 2025-01-20 & update & +``Lambda rejected (cold starts)''; ``EKS rejected (complexity)'' \\
2 & 2025-05-15 & update & +``12 Fargate services at 50K rps'' \\
\bottomrule
\end{tabular}
\caption{State evolution for c4-out-of-order. Note $s_1$ timestamp is older than $s_0$.}
\end{table}

\textbf{Key challenge:} State 1 has timestamp \texttt{2025-01-20}, which is \emph{older} than state 0's \texttt{2025-03-10}. The system ingests the decision document first, then receives an earlier evaluation document. The temporal guard should prevent the older document from overwriting the decision; instead, the system should \emph{enrich} the fact with the evaluation context (why Lambda and EKS were rejected).

\textbf{What intrinsic reveals:} If the system overwrites ``chose ECS Fargate'' with ``evaluating Lambda, ECS Fargate, and EKS'' (a regression), intrinsic scoring detects the loss of the decision claim. If it correctly enriches, all 4 claims from $C_1$ are present. Extrinsic QA may still pass either way if the LLM reasons over both fragments.

\section{System Prompts}
\label{app:prompts}

\subsection{Observation Extraction}

The prompts below are generalized for reproducibility. Unlike the other prompts in this section, which are domain-agnostic, this prompt is grounded in
the benchmark's business/customer-relationship domain, since the extraction categories define
what the system treats as a fact-worthy observation. The \texttt{\{categories\}} placeholder below
stands for the domain-specific observation categories actually used (e.g.\ a decision, a
commitment, an action item). Each extracted observation is further structured into
\texttt{who}/\texttt{what}/\texttt{when}/\texttt{why}/\texttt{where} fields; the statement referred
to below is the \texttt{what} field.

\begin{scriptsize}
\begin{verbatim}
You are analyzing text from a multi-author artifact.

Extract observations and entities from the text.

Observation guidelines:
- Extract every distinct piece of domain-relevant
  information.
- Each observation must contain exactly ONE factual
  claim.
- Every observation must be directly supported by
  the source text.
- Each observation's statement must be self-contained
  and understandable without the source document.
- Resolve all pronouns and references.
- Include temporal grounding when dates are available.
- Categories: {categories}
- Confidence: "High" = explicitly stated;
  "Medium" = discussed but not finalized;
  "Low" = inferred.

Entity guidelines:
- Extract all people mentioned by name, using the
  most complete form of each name.
- Extract other domain-relevant entities: {entity_types}
\end{verbatim}
\end{scriptsize}

\subsection{Fact Synthesis}

\begin{scriptsize}
\begin{verbatim}
You are synthesizing knowledge from observations
into evolving facts.

Decide how the new observation relates to existing
facts:
- "create": New knowledge not covered by any fact.
- "update": Adds new details to an existing fact.
  MERGE new information with existing.
- "reinforce": Confirms without adding new info.
- "supersede": Contradicts an existing fact.
- "none": Purely procedural, no lasting value.

Critical rules for "update":
- MUST preserve all details from existing statement.
- ADD new information (metrics, timelines, people).
- PREFER updating over creating.

Almost every observation matching an existing fact
should be "update." If in doubt, use "update."

Input:
New observation: {observation.what}
Existing facts for this topic: {existing_facts}
\end{verbatim}
\end{scriptsize}

\section{Evaluation Prompts}

\subsection{Claim Entailment Judge}

Given the system's active facts and an expected claim, determines whether the claim is entailed.

\begin{scriptsize}
\begin{verbatim}
Given these facts from a knowledge base:
1. {fact_1}
2. {fact_2}
...

Claim to verify: "{expected_claim}"

Which fact (if any) best matches or contains the
information in this claim?

Reply in this exact format:
FACT_NUMBER: <number or 0 if no fact matches>
SCORE: <YES if clearly states the claim,
        PARTIAL if related but incomplete,
        NO if no fact matches>
\end{verbatim}
\end{scriptsize}

Scoring: YES = 1.0, PARTIAL = 0.5, NO = 0.0

\subsection{Answer Generation (Extrinsic)}

Asks an agent using the memory system, given retrieved facts and a diagnostic question, to generate an answer.

\begin{scriptsize}
\begin{verbatim}
Answer the following question based ONLY on the
facts provided. If the facts don't contain enough
information, say "I don't have that information."

Facts:
1. {retrieved_fact_1}
2. {retrieved_fact_2}
...

Question: {extrinsicQA_question}

Answer concisely.
\end{verbatim}
\end{scriptsize}

\subsection{Answer Judge (Extrinsic)}

Compares the generated answer against the expected answer.

\begin{scriptsize}
\begin{verbatim}
Does the generated answer contain the same
information as the expected answer?

Generated answer: "{generated_answer}"
Expected answer: "{expected_answer}"

Answer YES if the generated answer conveys the
same key information.
Answer PARTIAL if it contains some but not all
key information.
Answer NO if it contradicts or misses the expected
information entirely.

Answer with exactly one word: YES, PARTIAL, or NO.
\end{verbatim}
\end{scriptsize}

Scoring: YES = 1.0, PARTIAL = 0.5, NO = 0.0

\section{Models Used}
\label{app:models}

Table~\ref{tab:models} lists all models used in the system under test and the evaluation scorer. The system uses a single frontier LLM for both observation extraction (converting artifact text into atomic observations) and fact synthesis (classifying the relationship between new observations and existing facts, then rewriting the fact statement). Titan Embeddings V2~\cite{titanembedv2} generates 1024-dimensional vectors for semantic search. Retrieval combines BM25~\cite{robertson2009bm25} keyword matching (weight 0.3) with k-NN vector similarity (weight 0.7). The evaluation scorer uses Claude Haiku 4.5~\cite{claudehaiku45} for judging and Amazon Nova Pro~\cite{novatechreport2025} for answer generation. All LLM calls use temperature 0 for deterministic outputs. All models are accessed via cloud-hosted APIs.

\begin{table}[h]
\centering
\scriptsize
\begin{tabular}{@{}llc@{}}
\toprule
\textbf{Component} & \textbf{Model} & \textbf{Temp.} \\
\midrule
\multicolumn{3}{@{}l}{\textit{System under test}} \\
Observation extraction & Frontier LLM & 0 \\
Fact synthesis & Frontier LLM & 0 \\
Embedding & Titan Embeddings V2 (1024-dim) & -- \\
Retrieval & BM25 (0.3) + k-NN (0.7) & -- \\
\midrule
\multicolumn{3}{@{}l}{\textit{Evaluation scorer}} \\
Claim entailment judge & Claude Haiku 4.5 & 0 \\
Answer generation & Amazon Nova Pro & 0 \\
Answer judge & Claude Haiku 4.5 & 0 \\
Embedding (claim matching) & Titan Embeddings V2 (1024-dim) & -- \\
\bottomrule
\end{tabular}
\caption{All models used in the system and evaluation pipeline.}
\label{tab:models}
\end{table}

\section{Per-State Coverage Degradation}

Table~\ref{tab:perstate} shows how claim coverage degrades as more documents are ingested within a trajectory. Later states require the system to retain earlier claims while incorporating new information, testing memory persistence.

\begin{table}[h]
\centering
\scriptsize
\begin{tabular}{@{}lcccc@{}}
\toprule
\textbf{State} & \textbf{All (\%)} & \textbf{L1 (\%)} & \textbf{L2 (\%)} & \textbf{L3 (\%)} \\
\midrule
$s_0$ (n=330) & 64.7{\scriptsize$\pm$28.8} & 67.5{\scriptsize$\pm$24.4} & 76.4{\scriptsize$\pm$22.1} & 50.2{\scriptsize$\pm$32.4} \\
$s_1$ (n=330) & 65.3{\scriptsize$\pm$22.1} & 72.7{\scriptsize$\pm$23.6} & 70.3{\scriptsize$\pm$20.3} & 52.9{\scriptsize$\pm$16.0} \\
$s_2$ (n=319) & 58.2{\scriptsize$\pm$20.8} & 68.2{\scriptsize$\pm$23.0} & 60.6{\scriptsize$\pm$17.9} & 46.1{\scriptsize$\pm$13.4} \\
$s_3$ (n=242) & 57.8{\scriptsize$\pm$18.8} & 62.3{\scriptsize$\pm$27.1} & 68.4{\scriptsize$\pm$14.8} & 48.2{\scriptsize$\pm$8.6} \\
$s_4$ (n=110) & 49.7{\scriptsize$\pm$11.4} & -- & 57.3{\scriptsize$\pm$10.9} & 47.8{\scriptsize$\pm$10.7} \\
$s_5$ (n=44) & 50.8{\scriptsize$\pm$10.8} & -- & -- & 50.8{\scriptsize$\pm$10.8} \\
$s_6$ (n=22) & 52.3{\scriptsize$\pm$10.4} & -- & -- & 52.3{\scriptsize$\pm$10.4} \\
\bottomrule
\end{tabular}
\caption{Claim coverage (\%) by state index, averaged across all trajectories and $N{=}11$ runs. $n$ = number of trajectory$\times$run pairs that reach this state (not all trajectories have the same number of states). Coverage decreases at later states as the system must retain earlier claims while integrating new information. L3 trajectories start lower and degrade further.}
\label{tab:perstate}
\end{table}

Coverage drops from 65\% at $s_0$ to 50\% at $s_4$, confirming that memory retention degrades as trajectories progress. At L2, the drop is steepest between $s_1$ (70\%) and $s_2$ (61\%), suggesting the system struggles to retain claims once 3+ documents have been processed. L3 starts lowest (50\% at $s_0$) due to the extraction challenge of hybrid-injection documents.

Note that \emph{fact coverage} (whether the system creates any fact related to the trajectory topic) is 98.8\% overall , the system almost always creates a fact. The low \emph{claim coverage} reported here reflects that the created facts are incomplete or fragmented, not that they are entirely absent.

Trajectory length (number of states) is moderately correlated with lower intrinsic accuracy (Spearman $\rho = -0.45$, $p < 0.02$): trajectories with 2--3 states average $\sim$71\% accuracy while those with 5--7 states average $\sim$49\%. This confirms that longer fact evolution sequences are harder to track, independently of the difficulty level assignment.

\section{Update Detection, False Updates, and Confidence Calibration}
\label{app:udrfur}

Table~\ref{tab:udr_fur} shows update detection recall (UDR), false update rate (FUR), and confidence MAE per category and difficulty, all averaged across $N{=}11$ runs.

\begin{table}[h]
\centering
\scriptsize
\begin{tabular}{@{}lccc@{}}
\toprule
\textbf{Category} & \textbf{UDR (\%)} & \textbf{FUR (\%)} & \textbf{Conf. MAE} \\
\midrule
C1: LLM Comp. & 51.4{\scriptsize$\pm$6.0} & 39.4{\scriptsize$\pm$1.9} & 0.105{\scriptsize$\pm$0.005} \\
C2: Retrieval & 94.8{\scriptsize$\pm$4.3} & 0.0{\scriptsize$\pm$0.0} & 0.092{\scriptsize$\pm$0.009} \\
C3: Classif. & 70.5{\scriptsize$\pm$9.6} & 45.5{\scriptsize$\pm$6.0} & 0.117{\scriptsize$\pm$0.007} \\
C4: Synthesis & 81.4{\scriptsize$\pm$3.3} & 15.7{\scriptsize$\pm$2.1} & 0.097{\scriptsize$\pm$0.006} \\
C5: Temporal & 90.6{\scriptsize$\pm$2.2} & 0.0{\scriptsize$\pm$0.0} & 0.108{\scriptsize$\pm$0.007} \\
C6: Multi-Auth. & 50.8{\scriptsize$\pm$7.4} & 38.6{\scriptsize$\pm$3.1} & 0.124{\scriptsize$\pm$0.005} \\
\midrule
\textbf{Difficulty} & & & \\
\midrule
L1 & 55.2{\scriptsize$\pm$3.9} & 37.4{\scriptsize$\pm$2.4} & 0.089{\scriptsize$\pm$0.003} \\
L2 & 81.6{\scriptsize$\pm$5.3} & 19.4{\scriptsize$\pm$1.3} & 0.112{\scriptsize$\pm$0.003} \\
L3 & 84.1{\scriptsize$\pm$2.1} & 9.8{\scriptsize$\pm$0.7} & 0.118{\scriptsize$\pm$0.004} \\
\midrule
\textbf{Overall} & 73.6{\scriptsize$\pm$2.6} & 22.2{\scriptsize$\pm$1.2} & 0.106{\scriptsize$\pm$0.002} \\
\bottomrule
\end{tabular}
\caption{Update detection recall (UDR), false update rate (FUR), and confidence MAE per category and difficulty, averaged across $N{=}11$ runs ($\pm$ std across runs). C2 and C5 never produce false updates but C1 and C6 frequently misclassify reinforcements as updates. Confidence calibration is worst for C6 (multi-author) and best for C4 (synthesis). FUR decreases with difficulty because at L3 the system rarely modifies facts.}
\label{tab:udr_fur}
\end{table}

C1 and C6 show both low UDR ($\sim$51\%) and high FUR ($\sim$39\%): the system misses true updates while simultaneously making false updates, a double failure that intrinsic evaluation exposes but extrinsic cannot distinguish from simple low accuracy. C2 and C5 achieve near-perfect UDR with zero false updates, indicating the classification challenge is category-specific rather than a general system weakness.

Confidence calibration is relatively uniform across categories (range 0.092--0.124), with an overall MAE of 0.106 indicating systematic miscalibration of approximately 10 points. C6 (Multi-Author) shows the worst calibration (0.124), likely because the system does not adequately adjust confidence based on author diversity. Calibration degrades slightly with difficulty (0.089 at L1 to 0.118 at L3), consistent with the system being less certain about harder trajectories.

\section{Metric Correlations}

To validate that our intrinsic metrics capture meaningful signal, we compute correlations between claim miss rate, intrinsic accuracy, and extrinsic accuracy across all 30 trajectories (using per-trajectory means across $N{=}11$ runs).

\begin{table}[h]
\centering
\scriptsize
\begin{tabular}{@{}lcc@{}}
\toprule
\textbf{Comparison} & \textbf{Pearson $r$} & \textbf{$p$-value} \\
\midrule
Miss rate vs.\ Intrinsic accuracy & $-0.841$ & $5.9 \times 10^{-9}$ \\
Miss rate vs.\ Extrinsic accuracy & $-0.391$ & $3.2 \times 10^{-2}$ \\
Intrinsic vs.\ Extrinsic accuracy & $+0.552$ & $1.6 \times 10^{-3}$ \\
\bottomrule
\end{tabular}
\caption{Pearson correlations across 30 trajectories. Miss rate strongly predicts intrinsic accuracy ($r = -0.84$) but weakly predicts extrinsic accuracy ($r = -0.39$), confirming that intrinsic metrics capture signal that extrinsic evaluation misses. The moderate intrinsic-extrinsic correlation ($r = 0.55$) indicates the two methods share some signal but measure substantially different properties.}
\label{tab:correlations}
\end{table}

The strong negative correlation between miss rate and intrinsic accuracy ($r = -0.84$, $p < 10^{-8}$) validates that claim coverage is the primary driver of intrinsic scores. The weaker correlation between miss rate and extrinsic accuracy ($r = -0.39$) confirms that extrinsic evaluation is less sensitive to extraction failures , consistent with LLM compensation masking missing claims at query time. The moderate positive correlation between intrinsic and extrinsic ($r = 0.55$) shows they are related but far from equivalent, reinforcing the paper's argument that both are needed.

Table~\ref{tab:signed} shows the signed divergence (intrinsic $-$ extrinsic) per category. Positive values indicate intrinsic scores higher than extrinsic (memory underestimated by QA).

\begin{table}[h]
\centering
\scriptsize
\begin{tabular}{@{}lcc@{}}
\toprule
\textbf{Category} & \textbf{Signed Div.} & \textbf{Interpretation} \\
\midrule
C1: LLM Comp. & $+0.276 \pm 0.260$ & Memory much better than QA reveals \\
C2: Retrieval & $+0.094 \pm 0.137$ & Memory slightly better \\
C3: Classif. & $+0.220 \pm 0.276$ & Memory much better than QA reveals \\
C4: Synthesis & $+0.036 \pm 0.173$ & Nearly neutral \\
C5: Temporal & $+0.120 \pm 0.160$ & Memory moderately better \\
C6: Multi-Auth. & $-0.003 \pm 0.158$ & No directional bias \\
\bottomrule
\end{tabular}
\caption{Signed divergence (intrinsic $-$ extrinsic) per category. C1 shows strong positive bias by design: memory is correct but QA underestimates. C6 shows no directional bias: both metrics are equally poor at capturing multi-author attribution.}
\label{tab:signed}
\end{table}

\section{Robustness Across Systems: Configuration Comparison}
\label{app:robustness}

Table~\ref{tab:robustness} reports headline statistics for the four configurations discussed in
Finding~5 of the main paper. All values are averaged across $N{=}11$ runs. The four configurations
are:
\begin{itemize}
  \item \textbf{Simpler sys.}: a simpler configuration of the memory system, on the pre-release corpus.
  \item \textbf{Main sys.}: the main system configuration, on the same pre-release corpus (differs
  from Simpler sys.\ only in the system).
  \item \textbf{Main (released)}: the main system configuration, on the publicly released
  corpus used for all other results in this paper (differs from Main sys.\ only in the corpus).
  \item \textbf{Upgraded backbone}: the main system configuration with a newer-generation frontier
  model as the backbone, replacing the one used in all other configurations, on the publicly
  released corpus.
\end{itemize}
The intrinsic--extrinsic divergence holds across all four: positive signed divergence,
near-chance pass/fail agreement, and MAD and DivRate in a narrow range, indicating the finding is
not an artifact of a single system implementation, corpus version, or backbone model.

\begin{table*}[t]
  \centering
  \begin{tabular}{@{}lcccc@{}}
      \toprule
      \textbf{Metric} & \textbf{Simpler sys.} & \textbf{Main sys.} & \textbf{Main (released)} & \textbf{Upgraded backbone} \\
      \midrule
      Intrinsic (\%)      & $57.6 \pm 1.8$ & $61.1 \pm 1.6$ & $60.6 \pm 1.6$ & $61.4 \pm 1.3$ \\
      Extrinsic (\%)      & $48.8 \pm 2.7$ & $48.5 \pm 2.1$ & $48.8 \pm 2.9$ & $49.2 \pm 3.4$ \\
      MAD                 & $0.164 \pm 0.027$ & $0.181 \pm 0.022$ & $0.168 \pm 0.025$ & $0.158 \pm 0.035$ \\
      DivRate (\%)        & $43.3 \pm 6.0$ & $49.4 \pm 7.2$ & $43.6 \pm 6.1$ & $39.7 \pm 6.1$ \\
      Signed divergence   & $+0.088 \pm 0.034$ & $+0.126 \pm 0.025$ & $+0.118 \pm 0.029$ & $+0.121 \pm 0.036$ \\
      Cohen's $\kappa$    & $0.104$ & $0.082$ & $0.118$ & $0.035$ \\
      \bottomrule
    \end{tabular}
  \caption{Headline statistics across the four configurations of Finding~5, each averaged over
    $N{=}11$ runs. \textbf{Main (released)} is the configuration used for all other results in this
    paper. Positive signed divergence and near-chance $\kappa$ persist in every configuration.}
  \label{tab:robustness}
\end{table*}

Beyond these aggregate statistics, the paper's category-level findings are also preserved under
the upgraded backbone. Comparing per-category metrics between the main configuration and the
upgraded backbone, the category rankings are highly stable: Spearman rank correlation across the
six categories is $\rho = 0.94$ for MAD, $\rho = 0.89$ for intrinsic accuracy, and $\rho = 0.94$
for classification accuracy. The named extremes the paper relies on are unchanged: C1 (LLM
Compensation) has the highest MAD and the highest classification accuracy, C6 (Multi-Author) has
the lowest intrinsic accuracy, and C5 (Temporal) has the lowest classification accuracy, in both
configurations. The only differences are adjacent swaps among already-close values, never a
reordering of these extremes, indicating the diagnostic category structure is not an artifact of
a particular backbone model.

\section{Per-Trajectory Results}
\label{app:pertraj}

Table~\ref{tab:pertraj} shows individual trajectory scores averaged across $N{=}11$ runs. Intr. = fact accuracy (Eq.~2 in the main paper): mean claim coverage across all states of the trajectory. Across runs, intrinsic accuracy ranges from 56.9\% to 63.3\% (std = 1.6\%) and extrinsic from 45.0\% to 55.0\% (std = 2.9\%), confirming high reproducibility despite LLM non-determinism. Intrinsic scoring is more stable than extrinsic because it evaluates the persistent store directly rather than depending on retrieval + generation.

\begin{table*}[t]
\centering
\small
\begin{tabular}{@{}lllccccccc@{}}
\toprule
\textbf{Trajectory} & \textbf{Cat} & \textbf{Diff} & \textbf{Intr. (\%)} & \textbf{Extr. (\%)} & \textbf{MAD} & \textbf{Miss (\%)} & \textbf{FC} & \textbf{CA (\%)} & \textbf{Dir} \\
\midrule
c1-confidence-001 & C1 & L1 & 100.0{\small$\pm$0.0} & 95.5{\scriptsize$\pm$14.4} & .045 & 11{\scriptsize$\pm$21} & 6.4{\scriptsize$\pm$4.8} & 100{\scriptsize$\pm$0} & $=$ \\
c1-duplicate-facts-001 & C1 & L1 & 100.0{\small$\pm$0.0} & 50.0{\scriptsize$\pm$0.0} & .500 & 17{\scriptsize$\pm$5} & 8.0{\scriptsize$\pm$3.7} & 100{\scriptsize$\pm$0} & $\uparrow$ \\
c1-stale-state-001 & C1 & L2 & 83.3{\small$\pm$12.3} & 45.5{\scriptsize$\pm$33.4} & .470 & 69{\scriptsize$\pm$9} & 14.5{\scriptsize$\pm$8.2} & 0{\scriptsize$\pm$0} & $\uparrow$ \\
c1-wrong-provenance-001 & C1 & L2 & 79.1{\small$\pm$5.1} & 45.5{\scriptsize$\pm$14.4} & .336 & 34{\scriptsize$\pm$14} & 21.3{\scriptsize$\pm$5.0} & 6{\scriptsize$\pm$13} & $\uparrow$ \\
c1-unresolved-conflict-001 & C1 & L3 & 48.5{\small$\pm$4.8} & 36.4{\scriptsize$\pm$22.3} & .121 & 99{\scriptsize$\pm$4} & 15.8{\scriptsize$\pm$5.0} & 2{\scriptsize$\pm$7} & $\uparrow$ \\
\midrule
c2-phrasing-mismatch-001 & C2 & L1 & 76.1{\small$\pm$3.6} & 50.0{\scriptsize$\pm$0.0} & .261 & 31{\scriptsize$\pm$3} & 10.6{\scriptsize$\pm$5.0} & 0{\scriptsize$\pm$0} & $\uparrow$ \\
c2-temporal-mismatch-001 & C2 & L1 & 59.1{\small$\pm$13.0} & 50.0{\scriptsize$\pm$0.0} & .121 & 60{\scriptsize$\pm$15} & 10.8{\scriptsize$\pm$4.5} & 3{\scriptsize$\pm$10} & $\uparrow$ \\
c2-buried-fact-001 & C2 & L2 & 61.8{\small$\pm$3.9} & 54.5{\scriptsize$\pm$14.4} & .127 & 67{\scriptsize$\pm$7} & 13.6{\scriptsize$\pm$7.7} & 0{\scriptsize$\pm$0} & $\uparrow$ \\
c2-compound-narrow-001 & C2 & L3 & 45.5{\small$\pm$9.0} & 50.0{\scriptsize$\pm$0.0} & .076 & 89{\scriptsize$\pm$7} & 24.0{\scriptsize$\pm$9.8} & 20{\scriptsize$\pm$23} & $=$ \\
c2-multi-fact-scatter-001 & C2 & L3 & 59.1{\small$\pm$6.6} & 50.0{\scriptsize$\pm$0.0} & .091 & 84{\scriptsize$\pm$6} & 31.9{\scriptsize$\pm$17.0} & 0{\scriptsize$\pm$0} & $\uparrow$ \\
\midrule
c3-reinforce-001 & C3 & L1 & 84.1{\small$\pm$12.0} & 59.1{\scriptsize$\pm$19.3} & .250 & 32{\scriptsize$\pm$15} & 10.8{\scriptsize$\pm$4.4} & 68{\scriptsize$\pm$44} & $\uparrow$ \\
c3-entity-split-001 & C3 & L2 & 90.9{\small$\pm$19.3} & 63.6{\scriptsize$\pm$22.3} & .273 & 30{\scriptsize$\pm$24} & 10.3{\scriptsize$\pm$5.6} & 0{\scriptsize$\pm$0} & $\uparrow$ \\
c3-subtle-update-001 & C3 & L2 & 59.1{\small$\pm$28.7} & 27.3{\scriptsize$\pm$44.5} & .409 & 24{\scriptsize$\pm$22} & 10.9{\scriptsize$\pm$7.8} & 55{\scriptsize$\pm$14} & $\uparrow$ \\
c3-contradiction-001 & C3 & L3 & 49.2{\small$\pm$6.6} & 45.5{\scriptsize$\pm$14.4} & .098 & 87{\scriptsize$\pm$8} & 22.9{\scriptsize$\pm$13.5} & 0{\scriptsize$\pm$0} & $=$ \\
\midrule
c4-simple-merge-001 & C4 & L1 & 50.0{\small$\pm$7.1} & 50.0{\scriptsize$\pm$0.0} & .030 & 95{\scriptsize$\pm$15} & 7.2{\scriptsize$\pm$5.7} & 0{\scriptsize$\pm$0} & $=$ \\
c4-dedup-001 & C4 & L2 & 54.5{\small$\pm$14.4} & 50.0{\scriptsize$\pm$0.0} & .045 & 98{\scriptsize$\pm$0} & 14.8{\scriptsize$\pm$9.4} & 100{\scriptsize$\pm$0} & $=$ \\
c4-out-of-order-001 & C4 & L2 & 48.2{\small$\pm$5.7} & 50.0{\scriptsize$\pm$0.0} & .036 & 60{\scriptsize$\pm$7} & 6.1{\scriptsize$\pm$3.4} & 5{\scriptsize$\pm$14} & $=$ \\
c4-scope-expansion-001 & C4 & L2 & 78.4{\small$\pm$7.7} & 45.5{\scriptsize$\pm$14.4} & .330 & 34{\scriptsize$\pm$15} & 19.1{\scriptsize$\pm$7.5} & 0{\scriptsize$\pm$0} & $\uparrow$ \\
c4-compound-001 & C4 & L3 & 37.5{\small$\pm$0.0} & 50.0{\scriptsize$\pm$0.0} & .125 & 100{\scriptsize$\pm$0} & 19.1{\scriptsize$\pm$6.8} & 0{\scriptsize$\pm$0} & $\downarrow$ \\
c4-noise-filter-001 & C4 & L3 & 48.2{\small$\pm$8.3} & 50.0{\scriptsize$\pm$0.0} & .055 & 72{\scriptsize$\pm$3} & 18.3{\scriptsize$\pm$7.4} & 0{\scriptsize$\pm$0} & $=$ \\
\midrule
c5-delayed-update-001 & C5 & L1 & 62.1{\small$\pm$7.4} & 40.9{\scriptsize$\pm$19.3} & .212 & 60{\scriptsize$\pm$6} & 12.6{\scriptsize$\pm$6.7} & 0{\scriptsize$\pm$0} & $\uparrow$ \\
c5-linear-001 & C5 & L1 & 64.5{\small$\pm$15.6} & 50.0{\scriptsize$\pm$0.0} & .164 & 55{\scriptsize$\pm$19} & 12.4{\scriptsize$\pm$5.5} & 0{\scriptsize$\pm$0} & $\uparrow$ \\
c5-retraction-001 & C5 & L2 & 64.5{\small$\pm$11.6} & 50.0{\scriptsize$\pm$0.0} & .164 & 60{\scriptsize$\pm$6} & 16.0{\scriptsize$\pm$7.8} & 0{\scriptsize$\pm$0} & $\uparrow$ \\
c5-gradual-drift-001 & C5 & L3 & 53.0{\small$\pm$13.9} & 50.0{\scriptsize$\pm$0.0} & .091 & 81{\scriptsize$\pm$8} & 24.4{\scriptsize$\pm$14.8} & 2{\scriptsize$\pm$6} & $=$ \\
c5-scope-narrowing-001 & C5 & L3 & 52.3{\small$\pm$8.9} & 45.5{\scriptsize$\pm$14.4} & .091 & 85{\scriptsize$\pm$11} & 33.1{\scriptsize$\pm$14.5} & 0{\scriptsize$\pm$0} & $\uparrow$ \\
\midrule
c6-authority-override-001 & C6 & L1 & 25.8{\small$\pm$20.6} & 13.6{\scriptsize$\pm$22.3} & .121 & 97{\scriptsize$\pm$7} & 6.4{\scriptsize$\pm$3.4} & 3{\scriptsize$\pm$10} & $\uparrow$ \\
c6-corroboration-001 & C6 & L1 & 58.0{\small$\pm$6.0} & 50.0{\scriptsize$\pm$0.0} & .080 & 73{\scriptsize$\pm$14} & 9.9{\scriptsize$\pm$4.8} & 64{\scriptsize$\pm$37} & $\uparrow$ \\
c6-disagreement-001 & C6 & L2 & 50.0{\small$\pm$0.0} & 45.5{\scriptsize$\pm$14.4} & .045 & 97{\scriptsize$\pm$0} & 14.7{\scriptsize$\pm$4.1} & 0{\scriptsize$\pm$0} & $=$ \\
c6-cross-role-synthesis-001 & C6 & L3 & 37.5{\small$\pm$11.9} & 50.0{\scriptsize$\pm$0.0} & .148 & 84{\scriptsize$\pm$7} & 15.7{\scriptsize$\pm$12.6} & 0{\scriptsize$\pm$0} & $\downarrow$ \\
c6-repetition-001 & C6 & L3 & 36.4{\small$\pm$6.4} & 50.0{\scriptsize$\pm$0.0} & .136 & 85{\scriptsize$\pm$9} & 11.3{\scriptsize$\pm$5.4} & 77{\scriptsize$\pm$27} & $\downarrow$ \\
\bottomrule
\end{tabular}
\caption{Per-trajectory results averaged across $N{=}11$ runs ($\pm$ std). MAD = mean absolute divergence between intrinsic and extrinsic for that trajectory. Miss = claim miss rate (\%, complement of claim coverage Eq.~1 in the main paper, computed across all claims in all states). FC = average fact count. CA = classification accuracy (\%). Dir = direction of divergence: $\uparrow$ = intrinsic $>$ extrinsic (memory underestimated), $\downarrow$ = extrinsic $>$ intrinsic (LLM compensates), $=$ = gap $<$ 5 points.}
\label{tab:pertraj}
\end{table*}

\end{document}